\def\year{2019}\relax
\documentclass[lette rpaper]{article} %DO NOT CHANGE THIS
\usepackage{aaai19}  %Required
\usepackage{times}  %Required
\usepackage{helvet}  %Required
\usepackage{courier}  %Required
\usepackage{url}  %Required
\usepackage{graphicx}  %Required
\usepackage{amsfonts,amssymb}
\usepackage{amsmath,bm,times, amsthm}
\usepackage{tikz}
\usepackage{graphicx}
\usepackage{algorithm}
\usepackage{algpseudocode}
\usepackage{color}
\usepackage{setspace}
\usepackage{booktabs}
\usepackage{array}
\usepackage{multirow}

\newtheorem{lemma}{\textit{\textbf{Lemma}}}
\newtheorem{lemmaproof}{\textit{\textbf{Proof}}}
\newtheorem*{lemmas}{\textit{\textbf{Lemma}}}
\newtheorem*{lemmaproofs}{\textit{\textbf{Proof}}}
\newtheorem*{claim}{\textit{\textbf{Claim}}}
\newtheorem*{claimproof}{\textit{\textbf{Proof}}}
\usetikzlibrary{shapes,arrows,shadows,decorations.pathreplacing,decorations.pathmorphing}
\begin{document}
%%%%  tikz  %%%%%%
% Define the layers to draw the diagram
\pgfdeclarelayer{background}
%\pgfdeclarelayer{middleground}
\pgfdeclarelayer{foreground}
\pgfsetlayers{background,main,foreground}

%%%%% Define block styles used later %%%%%%%%
%  adamcell
\tikzstyle{operation} = [rectangle, draw=black, fill=orange!20, text centered, minimum height = 3em, minimum width=3em ]
\tikzstyle{operation-legend} = [rectangle, draw=black, fill=orange!20, minimum height = 1em, minimum width=1.5em ]
\tikzstyle{nonlinear}=[circle, draw=black, fill=red!20]

\tikzstyle{norm-adam}=[rectangle, draw=black, fill=olive!20, text centered, minimum height = 2em, minimum width=3em ]

\tikzstyle{cell} = [draw=blue!50, fill=blue!20, text width=27em, 
text centered, minimum height=16em,rounded corners]

% framework
\tikzstyle{loss}=[rectangle, draw=red!50, fill=red!40, text centered, inner sep= 0pt, minimum size = 2.3em]
\tikzstyle{opt}=[rectangle, draw=blue!100, fill=blue!20, text centered, inner sep= 0pt, minimum size = 2em]
\tikzstyle{add} = [circle, draw=orange!50, fill=orange!20]

\tikzstyle{block1}=[draw=gray!60,text width=8.2em, 
text centered, minimum height=2.2em,rounded corners]
\tikzstyle{block2}=[draw=purple!50,text width=8.2em, 
text centered, minimum height=2.2em,rounded corners]
\tikzstyle{block3}=[draw=purple!100,text width=8.2em, 
text centered, minimum height=2.2em,rounded corners]
\tikzstyle{block5}=[draw=olive!80,text width=8em, 
text centered, minimum height=2.2em,rounded corners]
\tikzstyle{block6}=[draw=darkgray!80,text width=8.2em, 
text centered, minimum height=2.2em,rounded corners]
\tikzstyle{block4}=[draw=orange!80,dashed,minimum width=9.6em, minimum height=6.4em, rounded corners,thick]

\tikzstyle{vectorh3} = [circle,thick, draw=teal!100, minimum size = 2em]
\tikzstyle{vectorhl} = [circle,minimum size = 2em]
\tikzstyle{vectorho} = [circle,minimum size = 1em]
\tikzstyle{operation1} = [rectangle, draw=teal!100,thick, fill=olive!50, text centered, inner sep= 0pt, minimum width = 9em, minimum height=2em]

% moment field
\tikzstyle{legend}=[rectangle, inner sep= 0pt, minimum height = 2.5em, text width=5em]

%statecell
\tikzstyle{vectorg} = [circle,thick, draw=teal!100, minimum size = 3em]
\tikzstyle{operation2} = [rectangle, draw=olive!50, thick, text centered, inner sep= 1pt, minimum width = 10em, minimum height=2.5em]
\tikzstyle{operation3} = [rectangle, draw=olive!50, thick, text centered, inner sep= 0pt, minimum width = 7em, minimum height=2.5em]
%%%%%%%%%%%%%%%%%%%%%%%%%%%%%%%%%%%%%%%%%%%%%%%%%%%%%%

\title{HyperAdam: A Learnable Task-Adaptive Adam for Network Training}
\author{Shipeng Wang \and Jian Sun\thanks{Corresponding author}  \and Zongben Xu\\
	School of Mathematics and Statistics, Xi'an Jiaotong University, Xi'an, 710049, China\\
	wangshipeng8128@stu.xjtu.edu.cn, \{jiansun, zbxu\}@xjtu.edu.cn 
}
\maketitle
\begin{abstract}
Deep neural networks are traditionally trained using human-designed stochastic optimization algorithms, such as SGD and Adam. Recently, the approach of learning to optimize network parameters has emerged as a promising research topic. However, these learned black-box optimizers sometimes do not fully utilize  the experience in human-designed optimizers, therefore have limitation in generalization ability. In this paper, a new optimizer, dubbed as \textit{HyperAdam}, is proposed that combines the idea of ``learning to optimize'' and traditional Adam optimizer. Given a network for training, its parameter update in each iteration generated by HyperAdam is an adaptive combination of multiple updates generated by Adam with varying decay rates. The combination weights and decay rates in HyperAdam are adaptively learned depending on the task.  HyperAdam is  modeled as a recurrent neural network with AdamCell, WeightCell and StateCell. It is justified to be state-of-the-art for various network training, such as multilayer perceptron, CNN and LSTM.
\end{abstract}
\section{Introduction}
Deep learning approach has exhibited strong capabilities in data representation \cite{lecun2015deep}, non-linear mapping \cite{sutskever2014sequence}, distribution learning \cite{goodfellow2014generative}, etc. Deep learning not only has wide applications in a broad field of academical studies, such as image analysis \cite{he2016deep}, speech recognition \cite{mcmahan2018listening}, robotics \cite{lillicrap2016continuous}, inverse problem \cite{sun2016deep}, but also draws attention of industry for realization in  products. 

One challenge in deep learning is the effective optimization of deep network parameters, required to be generalizable to varying network architectures, e.g., network type, depth, width, non-linear activation functions. For a neural network $f(x;w)$,  the aim of network training is to find the optimal network parameters $w^* \in \mathbb{R}^p$ to minimize empirical loss between the network output given input $x_i\in \mathbb{R}^d$ and the corresponding target label $y_i \in \mathbb{R}^b$:
\begin{equation} \label{empirical loss}
w^* = \arg \min_w \Sigma_{i=1}^N l(f(x_i;w),y_i),
\end{equation} 	
where $\{ (x_i, y_i)\}_{i=1}^N$ is the training set. For a deep neural network, the dimension $p$ and number of training data $N$  are commonly large in real applications. 

The network, as a learning machine, is referred to as a \textit{learner}, the loss for network training is defined as an \textit{optimizee}, and the optimization algorithm to minimize optimizee is referred to as an \textit{optimizer}. For example, a gradient-based optimizer can be written as a function $O$ that maps the gradient $g_t$ to network parameter update $d_t$ in $t$-th iteration:
\begin{equation}\label{opimizer}
d_t(\varTheta) = O(g_t, \mathcal{H}_{t}; \varTheta),
\end{equation}
where $\mathcal{H}_{t}$ represents the historical gradient information and $\varTheta$ represents the hyperparameters of the optimizer. 

The human-designed optimizers, such as stochastic gradient descent (SGD) \cite{robbins1951stochastic},  RMSProp \cite{tieleman2012lecture}, AdaGrad \cite{duchi2011adaptive}, AdaDelta \cite{zeiler2012adadelta} and Adam \cite{kingma2015adam}, are popular in network training. They have well generalization ability to various network architectures and tasks. Adam takes the statistics of gradients as the historical information recursively accumulated with constant decay rates (i.e., $\beta, \gamma$ in Alg. 1). Though universal, Adam suffers from unsatisfactory convergence in some cases because of the constant decay rates \cite{j2018on}.
				%suffers from slow convergence and difficulty in tuning hyperparameters similar to other traditional optimizers.

Recently, ``learning to optimize'', i.e., learning the optimizer by data-driven approach, triggered the interest of community. The optimizer \cite{andrychowicz2016learning} outputs the update vector by RNN, whose generalization ability is improved by two training tricks \cite{lv2017learning}. This idea is also applied to optimizing derivative-free black-box functions \cite{chen2017learning}. From the perspective of reinforcement learning, the optimizer is taken as policy \cite{li2016learning}. Though faster in decreasing training loss than the traditional optimizers in some cases, the learned optimizers do not always generalize well to diverse variants of learners. Moreover, they can not be guaranteed to output a descent direction in each iteration for network training. 

In this paper, we propose an effective optimizer \textit{HyperAdam}, which is a generalized and learnable optimizer inspired by Adam. For network optimization, the parameter update generated by HyperAdam is an ensemble of updates generated by Adam with different decay rates. Both decay rates and combination weights for ensemble are adaptively learned depending on the task. To implement this idea, AdamCell and WeightCell are respectively designed to generate candidate updates and weights to combine them, conditioned on the output of StateCell for modeling task-dependent state. As a recurrent neural network,  parameters of HyperAdam are learned by training on a meta-train set.

The main contribution of this paper is two-fold. First, to the best of our knowledge, this is a first task-adaptive optimizer taking  merits of adaptive moment estimation approach (i.e., Adam) and  learning-based approach in a single framework. It opens a new door to design learning-based optimizer inspired by traditional human-designed optimizers. Second,  extensive experiments justify that the learned HyperAdam outperforms traditional optimizers, such as Adam and learning-based optimizers for training a wide range of neural networks, e.g., deep MLP, CNN, LSTM.

\section{Related Works} 
\begin{algorithm}[!t]
	\caption{Adam Optimizer}
	\label{adam}
	\renewcommand{\algorithmicensure}{\textbf{Initialize:}}
	\begin{algorithmic}[1]
		\Require{~~\\
			Initialized parameter $w_0$,
			step size $\alpha$,
			batch size $N_B$.\\
			Exponential decay rates $\beta, \gamma$; dataset $\{ (x_i, y_i)\}_{i=1}^N$.
			%			Random batches$\{ (x_{t_k}, y_{t_k})\}_{k=1}^{N_t} (t=1,\dots,T)$ drawn from dataset $\{ (x_i, y_i)\}_{i=1}^N$
		}
		\Ensure {%~~\\
			$m_0=0,v_0=0$.
		}
		\ForAll {$t=1,\dots, T$}
		\State {Draw random batch $\{ (x_{i_k}, y_{i_k})\}_{k=1}^{N_B}$ from dataset}
		\State{$g_t =  \Sigma_{k=1}^{N_B} \nabla l(x_{i_k},y_{i_k}, w_{t-1})$}
		\State{$m_t = \beta m_{t-1} + (1-\beta)g_t$} \Comment{\textit{moving average}}
		\State{$v_t = \gamma v_{t-1} + (1-\gamma)g_t^2$}
		\State{$\tilde{m}_t = \frac{m_t}{1-\beta^t}$, $\tilde{v}_t = \frac{v_t}{1-\gamma^t}$} \Comment{\textit{correcting bias}}
		\State{$\hat{m}_t = \frac{\tilde{m}_t}{\sqrt{\tilde{v}_t} + \varepsilon}$}
		\State{$w_{t} = w_{t-1} - \alpha \hat{m}_t$}
		\EndFor	\\
		\Return{final parameter $w_{T}$.}
	\end{algorithmic}
\end{algorithm}

\begin{algorithm}[!t]
	\caption{Task-Adaptive HyperAdam}
	\label{generalized adam-optimzier}
	\renewcommand{\algorithmicensure}{\textbf{Initialize:}}
	\begin{algorithmic}[1]
		\Require{ ~~\\
			Initialized parameter $w_0$,
			step size $\alpha$,
			batch size $N_B$.\\
			%			Random batches$\{ (x_{t_k}, y_{t_k})\}_{k=1}^{N_t} (t=1,\dots,T)$ drawn from dataset $\{ (x_i, y_i)\}_{i=1}^N$;\\
			%Parametric functions: $F_h$, $F_u$, $F_r$, $F_q$;\\
			Dataset $\{ (x_i, y_i)\}_{i=1}^N$.
		}
		\Ensure{~~\\
			$\mathbf{m}_0$, $\mathbf{v}_0$, $\hat{\boldsymbol\beta}_0$,$ \hat{\boldsymbol\gamma}_0$, $\mathbf{s}_0$ $=\mathbf{0}\in\mathbb{R}^{p\times J}$,
			$ \mathbf{1}$ $\in\mathbb{R}^{p\times J}$,
			$\varepsilon$=1e-24
		}.
		\ForAll {$t=1,\dots,T$}
		\State {Draw random batch $\{ (x_{i_k}, y_{i_k})\}_{k=1}^{N_B}$ from dataset}
		\State{$g_t =  \Sigma_{k=1}^{N_B} \nabla l(x_{i_k},y_{i_k}, w_{t-1})$}
		\State{$\mathbf{G_t}=[g_t,\dots,g_t]$} \Comment{$\mathbf{G_t}\in\mathbb{R}^{p\times J}$}
		\State{$\mathbf{s}_t =F_h(\mathbf{s}_{t-1},g_t;\varTheta_h)$} \Comment{\textit{current state}}
		\State{{$\boldsymbol\beta_t\triangleq[\beta_t^1,\dots,\beta_t^J]=F_u(\mathbf{s}_t,\mathbf{m}_{t-1}; \varTheta_u)$}} %\textcolor{blue}
		\State{{$\boldsymbol\gamma_t\triangleq[\gamma_t^1,\dots,\gamma_t^J]=F_r(\mathbf{s}_t,\mathbf{m}_{t-1};\varTheta_r)$}}%\textcolor{blue}
		\State{{$\mathbf{m}_t=\boldsymbol{\beta}_t\odot \mathbf{m}_{t-1} + (\mathbf{1}-\boldsymbol{\beta}_t) \odot \mathbf{G_t}$}}
		\State{{$\mathbf{v}_t=\boldsymbol{\gamma}_t\odot \mathbf{v}_{t-1} + (\mathbf{1}-\boldsymbol{\gamma}_t) \odot \mathbf{G}_t^2$}}
		\State{{$\hat{\boldsymbol\beta}_{t} = \boldsymbol\beta_{t} \odot \hat{\boldsymbol\beta}_{t-1} + (\mathbf{1} - \boldsymbol\beta_{t}) \odot \mathbf{1}$}}
		\State{{$\hat{\boldsymbol\gamma}_{t} = \boldsymbol\gamma_{t} \odot \hat{\boldsymbol\gamma}_{t-1} + (\mathbf{1} - \boldsymbol\gamma_{t}) \odot \mathbf{1}$}}
%		\State {$\tilde{\mathbf{m}}_t = \frac{\mathbf{m}_t}{\hat{\boldsymbol\beta}_{t}}$, $\tilde{\mathbf{v}}_t = \frac{\mathbf{v}_t}{\hat{\boldsymbol\gamma}_{t}}$}, \Comment{\textit{correcting bias}}
		\State {$\tilde{\mathbf{m}}_t = {\mathbf{m}_t}/{\hat{\boldsymbol\beta}_{t}}$, $\tilde{\mathbf{v}}_t = {\mathbf{v}_t}/{\hat{\boldsymbol\gamma}_{t}}$}, \Comment{\textit{correcting bias}}
		\State{{$\hat{\mathbf{m}}_t \triangleq [\hat{m}_t^1,\dots,\hat{m}_t^J] =\frac{ \tilde{\mathbf{m}}_t }{\sqrt{\tilde{\mathbf{v}}_t  + \varepsilon}}$}} \Comment{\textit{moment field}}
		\State{$\boldsymbol{\rho}_t\triangleq[\rho_t^1,\dots,\rho_t^J]=F_q(\mathbf{s}_t;\varTheta_q)$} \Comment{\textit{weight field}}
		\State{$d_t = \Sigma_{j=1}^J \rho_t^j\odot \hat{m}_t^j$} %\Comment{$\hat{m}_t^j$ is the $j$-th column of $\hat{\mathbf{m}}_t$}
		\State{$w_{t} = w_{t-1} - \alpha d_t$}
		\EndFor	\\
		\Return{final parameter $w_{T}$.}
	\end{algorithmic}
\end{algorithm}

\begin{figure*}[!htbp]
	\centering
	
	\tikzset{global scale/.style={
			scale=#1,
			every node/.append style={scale=#1}
		}
	}
	\begin{tikzpicture}[global scale = 0.56]
	\begin{pgfonlayer}{main}
	% optimizee start
	\node[above](opt-zee){\LARGE Optimizee};
	% add 1
	\path (opt-zee.east) + (2.0, 0) node [add](add1){\LARGE $+$};
	\path [draw, ->] (opt-zee.east) -- (add1.west);
	% add 2
	\path (add1.east) + (2.0, 0) node [add](add2){\LARGE $+$};
	\path [draw, ->] (add1.east) -- (add2.west);
	% add 3
	\path (add2.east) + (2.0, 0) node [add](add3){\LARGE$+$};
	\path [draw,->] (add2.east) -- (add3.west);
	% optimizee end
	\path (add3.east) + (1.6, 0) node (opt-zee-end){};
	\path [draw,->] (add3.east) -- (opt-zee-end);
	% parameter t-1
	\path (opt-zee.east) + (0.7, 0) node[above] (para1) {\LARGE $w_{t-2}$};
	% parameter t
	\path (add1.east) + (0.7, 0) node [above] (para2){\LARGE$w_{t-1}$};
	% parameter t+1
	\path (add2.east) + (0.7, 0) node [above] (para3) {\LARGE$w_{t}$};
	% parameter t+2
	\path (add3.east) + (0.7, 0) node [above](para4) {\LARGE$w_{t+1}$};
	% loss t-1
	\path (opt-zee.east) + (1.2,1.5) node [loss](f1){\LARGE$L_{t-2}$};
	\path [draw,->] (opt-zee.center -| f1.south) -- (f1.south);
	% loss t
	\path (add1.east) + (1.2,1.5) node [loss](f2){\LARGE$L_{t-1}$};
	\path [draw,->] (add1.center -| f2.south) -- (f2.south);
	% loss t+1
	\path (add2.east) + (1.2,1.5) node [loss](f3){\LARGE$L_{t}$};
	\path [draw,->] (add2.center -| f3.south) -- (f3.south);
	% optimizer start
	\path (opt-zee.south) + (0,-3.0) node [above] (opt-zer) {\LARGE Optimizer};
	% optimizer 1
	\path (opt-zer.center -| add1.center) node [opt] (opt-zer1){\LARGE$O$};
	\path [draw,->] (opt-zer.east) -- (opt-zer1.west) node [midway, below]{\LARGE$\mathcal{H}_{t-2}$};
	% optimizer 2
	\path (opt-zer1.center -| add2.center) node [opt] (opt-zer2){\LARGE$O$};
	\path [draw,->] (opt-zer1.east) -- (opt-zer2.west) node [midway, below]{\LARGE$\mathcal{H}_{t-1}$};
	% optimizer 3
	\path (opt-zer.center -| add3.center) node [opt] (opt-zer3){\LARGE$O$};
	\path [draw,->] (opt-zer2.east) -- (opt-zer3.west)node [midway, below]{\LARGE$\mathcal{H}_{t}$};
	% optimizer end
	\path (opt-zer.center -| opt-zee-end.center) node (opt-zer-end){};
	\path [draw,->] (opt-zer3.east) -- (opt-zer-end.west) node [midway, below]{\LARGE$\mathcal{H}_{t+1}$};
	% update direction
	\path [draw, ->] (opt-zer1.north) -- (add1.south);
	\path (opt-zer1.north)+ (0, 1.6) node [right] {\LARGE$d_{t-1}$};
	\path [draw, ->] (opt-zer2.north) -- (add2.south) ; % node [midway,right] {$d_t$}
	\path (opt-zer2.north)+ (0, 1.6) node [right] {\LARGE$d_{t}$};
	\path [draw, ->] (opt-zer3.north) -- (add3.south) ; % node [midway,right] {$d_{t+1}$}
	\path (opt-zer3.north)+ (0, 1.6) node [right] {\LARGE$d_{t+1}$};
	% gradient
	\path [draw, dashed, ->] (para1.south) -- (opt-zer1.north -| opt-zer1.west);
	\path (para1.south) + (0.2, -1.7) node {\LARGE$g_{t-1}$};
	\path [draw, dashed, ->] (para2.south) -- (opt-zer2.north -| opt-zer2.west);
	\path (para2.south) + (0.2, -1.7) node {\LARGE$g_{t}$};
	\path [draw, dashed, ->] (para3.south) -- (opt-zer3.north -| opt-zer3.west);
	\path (para3.south) + (0.2, -1.7) node {\LARGE$g_{t+1}$};
	% optimizer
	% start
	\path (opt-zee-end) + (8, -3.5) node[vectorh3] (start) {\LARGE$g_t$};
	\path (start.north) + (0,0.8) node [operation1](lstm){\LARGE StateCell};
	\path [draw, thick,->] (start.north) --(lstm.south -| lstm.center);
	% level2 lstm
	\path (lstm.north) + (0,1) node[block1] (level2){};
	\path (level2.center) + (-2.8em, -0em) node[vectorhl]{\LARGE$s_{t}^1$};
	%	\path (level2.center) + (2.8em,0.5em) node(level22)[vectorhl] {};
	%	\draw [very thick,red!80,->] (level22.220)--(level22.40);
	\path (level2.center) + (2.8em, -0em) node[vectorhl]{\LARGE$s_{t}^J$};
	\path (level2.center) node[]{\LARGE$\dots$};	
	% lstm
	\path [draw,thick,->]  (lstm.north) -- (level2.south-|lstm.center) node [midway] (hid){};
	
	% level3 adam
	\path (level2.north) + (-5.5em,2) node[block5] (level31){};
	\path (level31.center) + (-2.8em, -0em) node[vectorho]{\LARGE$\hat{m}_{t}^1$};
	\path (level31.center) + (2.8em,-0em) node[vectorho] {\LARGE$\hat{m}_{t}^J$};
	\path (level31.center) node[]{$\dots$};
	% level3 fc elu
	\path (level2.north) + (5.5em,2) node[block6] (level32){};
	\path (level32.center) + (-2.8em, -0em) node[vectorho]{\LARGE$\rho_{t}^1$};
	\path (level32.center) + (2.9em, -0em) node[vectorho]{\LARGE$\rho_{t}^J$};
	\path (level32.center) node[]{\LARGE$\dots$};
	
	\path (level2.north) + (0,1.0) node (s23) {};
	\path [draw,thick] (level2.north) -- (s23);
	
	\path (level31.center |- s23.south) node [operation1] (adam)  {\LARGE AdamCell};
	\path [draw, thick, ->] (s23.south) -- (adam.east);
	\path [draw, thick, ->] (adam.north) -- (level31.south);
	
	\path (level32.center |- s23.south) node [operation1] (fcelu)  {\LARGE WeightCell};
	\path [draw, thick, ->] (s23.south) -- (fcelu.west);
	\path [draw, thick, ->] (fcelu.north) -- (level32.south);
	% level4
	\path (level31.north -| level2.center) + (0,2) node [vectorh3] (end) {\LARGE$d_t$};

	\path (end.south) + (0, -0.9) node [operation1] (lc) {\LARGE Combination};
	\path [draw, thick, ->] (level31.north) -- (level31.north |- lc.center) -- (lc.west);
	\path [draw, thick, ->] (lc.north) -- (end.south);
	\path [draw, thick, ->] (level32.north) -- (level32.north |- lc.center) -- (lc.east);
	
	% hidden state
	% ht	
	\path (lstm.west) + (-3.5, -0) node [block3] (h2t) {} ;
	\path (h2t.center) + (-2.8em,-0em) node[vectorho] {\LARGE$s_{t-1}^1$};
	\path (h2t.center) + (2.8em,-0em) node[vectorho] {\LARGE$s_{t-1}^J$};
	\path (h2t.center) node[]{\LARGE$\dots$};
	
	\path (h2t) + (0, 1) node [block2] (h1t) {};
	\path (h1t.center) + (-2.8em, -0.0em) node[vectorho]{\LARGE$m_{t-1}^1$};
	\path (h1t.center) + (2.8em, -0.0em) node[vectorho]{\LARGE$m_{t-1}^J$};
	\path (h1t.center) node[]{\LARGE$\dots$};
	
	\path (h2t.north) + (0, 0.3em) node [block4]{};
	
	%ht+1
	\path (lstm.east) + (3.5, -0) node [block3] (h2t+1) {} ;
	\path (h2t+1.center) + (-2.5em, -0em) node[vectorho]{\LARGE$s_{t}^1$};
	\path (h2t+1.center) + (2.5em, -0em) node[vectorho]{\LARGE$s^J_{t}$};
	\path (h2t+1.center) node[]{$\dots$};
	
	\path (h2t+1) + (0, 1) node [block2] (h1t+1) {};
	\path (h1t+1.center) + (-2.8em, -0em) node[vectorho]{\LARGE$m_{t}^1$};
	\path (h1t+1.center) + (2.8em, -0em) node[vectorho]{\LARGE$m_{t}^H$};
	\path (h1t+1.center) node[]{$\dots$};
	
	\path (h2t+1.north)+ (0, 0.3em) node [block4]{};
	
	% bn2
	\path [draw, blue!100,dashed, line width=1.0pt,->] (h1t.north)--(h1t.north|-adam.west)--(adam.west);
	\path (adam.south) + (0, -0.4) node (a) {};
	\path [draw, blue!100,dashed,line width=1.0pt, ->] (adam.south)--(a.north)--(a.north-|h1t+1.center) -- (h1t+1.north);
	\path (h2t.east) + (1,0) node (b) {};
	\path (hid.center) + (2.4,0) node(c) {};	
	\path [draw, blue!100,dashed,line width=1.0pt, ->] (h2t.east) -- (b.center)-- (b.center |- lstm.west)--(lstm.west);
	\path [draw, blue!100,dashed,line width=1.0pt, ->] (hid.center)--(c.center)--(c.center|-h2t+1.west) -- (h2t+1.west);
	
	% notation
	\path (h2t.south) + (0,-0.6) node {\LARGE$\mathcal{H}_{t-1}$};
	\path (h2t+1.south) + (0,-0.6) node {\LARGE$\mathcal{H}_{t}$};
	\path (level31.west) + (-1.6, -0) node {\LARGE Moment field};
	\path (level32.east) + (1.6, -0) node {\LARGE Weight field};
	
	% o
	\path (opt-zer3.south) + (2.1, -0.8) node (ctrl11){};
	\path (opt-zer3.south) + (2.2, -0.4) node (ctrl12) {};
	\path (opt-zer3.south) + (2.5, 0.5) node (end11) {} ;
	\draw [dotted,thick](opt-zer2.south) .. controls (ctrl11.center) and (ctrl12)..(end11.center);
	\path (opt-zer3.south) + (3.5, -0.44) node (end2) {};
	\draw [dotted,thick] (opt-zer2.south) to[bend right] (end2.center);

	\end{pgfonlayer}
	\end{tikzpicture}
	
	\caption{Computational graph of  HyperAdam. $O$ represents the optimizer. $g_t$ is the gradient of the optimizee $L$ and $d_t$ is the update vectors. The historical information $\mathcal{H}_{t}$ consists of the first moment and previous task state.}% For clarity, the cell state of AdamCell and StateCell are omitted.}
	\label{framework}
	
\end{figure*}
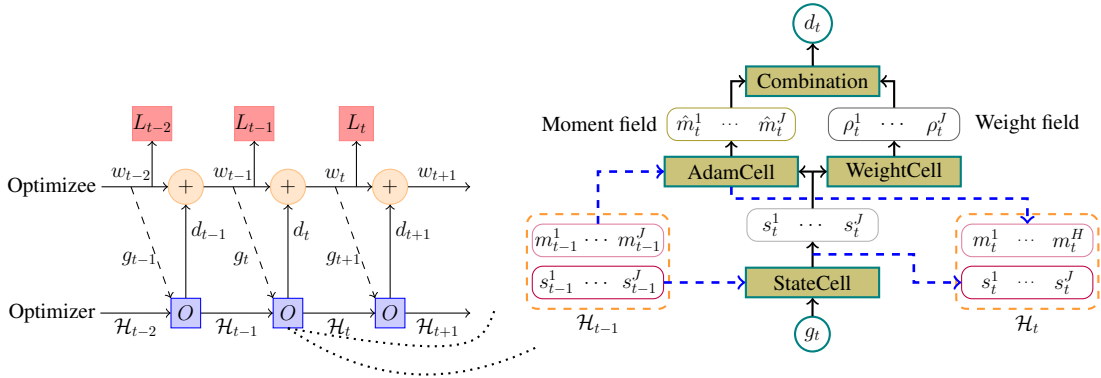 

\subsection{Learning to Optimize}
With the goal of facilitating learning of novel  tasks, \textit{meta-learning} is developed to extract knowledge from observed tasks \cite{amit2018priors,ren2018metalearning,finn2017one,snell2017prototypical,wichrowska2017learned,santoro2016meta,Daniel2016}.  %,andrychowicz2016learning 

This paper focuses on the meta-learning task of optimizing network parameters, commonly termed as ``learning to optimize". It originates from several decades ago \cite{schmidhuber1992learning,naik1992meta} and is developed afterwards \cite{hochreiter2001learning,younger2001meta}. Recently, a more general optimizer that conducts parameter update by LSTM with gradient as input is proposed in \cite{andrychowicz2016learning}. Two effective training techniques, ``Random Scaling'' and ``Combination with Convex Functions'', are proposed to improve the generalization ability \cite{lv2017learning}. Subsequently, several works use RNN to replace certain process in some optimization algorithms, e.g., variational EM \cite{marino2018learning}, ADMM \cite{liu2017proximal}. In \cite{chen2017learning}, RNN is also used to optimize derivate-free black-box functions.

These pioneering learning-based optimizers have shown promising performance, but did not fully utilize the experience in human-designed optimizers, and sometimes have limitation in generalizing to variants of networks. The proposed optimizer, HyperAdam, is a learnable optimizer but with architecture designed by generalizing traditional Adam optimizer. In the evaluation section, the HyperAdam is justified to have better generalization ability than previous learning-based optimizers for training various networks.

\subsection{Adam Method}
Vanilla SGD has been improved by adaptive learning rates for each parameter (e.g., AdaGrad, AdaDelta, RMSProp) or (and) Momentum \cite{tseng1998incremental}. Adam \cite{kingma2015adam} is an adaptive moment estimation method combining these two techniques, as illustrated in Alg. \ref{adam}. Adam takes unbiased estimation of second moment of gradients as the ingredient of the coordinate-wise learning rates, and the unbiased estimation of first moment of gradients as the basis for parameter updating. The bias is caused by the initialization of mean and uncentered variance during online moving average with decay rates (i.e., $\beta,\gamma$ in Alg.~\ref{adam}). It is easy to verify that the parameters update generated by Adam is invariant to the scale of gradients when ignoring $\varepsilon$. 

As observed in \cite{j2018on}, Adam suffers from unsatisfactory convergence due to the constant decay rates when the variance of gradients with respect to optimization steps are large. While, in HyperAdam, the generalized Adam are with learned decay rates adaptive to task state and gradients. Moreover, the ensemble technique  \cite{wolpert1992stacked} of HyperAdam that combines multiple candidate  updates can potentially find more reliable descent directions. These techniques are justified to be effective for improving the baseline Adam in evaluation section.
\section{HyperAdam}
In this section, we introduce the general idea, algorithm and network architecture of the proposed HyperAdam.
\begin{figure}[!bp]
	
	\centering
	\tikzset{global scale/.style={
			scale=#1,
			every node/.append style={scale=#1}
		}
	}
	\begin{tikzpicture}[global scale = 0.45]
	\path node (start) {};
	\path (start) + (0.2, 0.2) node {\Huge$w_0$};
	\path (start.center) + (1,-3) node (step1) {};
	\path [draw, ->, red, thick] (start.center) -- (step1.center) node [midway,right, black] {\Huge $d_1$};
	\path (step1.west) + (-0.20,-0.20) node {\Huge $w_1$};
	\path (step1.center) + (1, -0.6) node (step-t-1) {};
	\path [draw, dotted, thick, red] (step1.center) -- (step-t-1.center) {};
	\path (step-t-1.center) + (0.40,-0.30) node {\Huge $w_{t-1}$};
	\path (step-t-1.center) + (1.5, 2.5) node (step-t) {};
	\path (step-t.center) + (-0.40,0.30) node {\Huge $w_{t}$};
	\path [draw, ->, red, thick] (step-t-1.center) -- (step-t.center) node [midway,right] (dt-1)  {};
	\path (dt-1.center) + (0.1, -0.1) node[right] {\Huge $d_{t}$};   
	% Momentum 1
	\path (step-t.center) + (1.2,1) node [above](m-t-1) {};
	\path [draw, dashed, ->, blue, thick] (step-t.center)--(m-t-1.center);
	
	\path (step-t.center) + (1.6,0.8) node (m-t-2) {};
	\path [draw, dashed, ->, blue, thick] (step-t.center)--(m-t-2.center);
	\path (step-t.center) + (1.8,-0.3) node (m-t-4) {};
	\path (m-t-4.center) + (0.6, - 0.1) node (saddle) {};
	\fill [teal!80] (saddle.center) circle (4pt);
	\path (saddle.center) + (0.5,-0.7) node {\huge Saddle point};
	\path [draw, dashed, ->, blue, thick] (step-t.center)--(m-t-4.center);
	% step t
	\path (step-t.center) + (3, 1) node (step-t1){};
	\path [draw, ->, red,thick] (step-t.center) -- (step-t1.center) node [midway] (dt) {};
	\path (step-t1.center) + (-0.40,0.30) node {\Huge $w_{t+1}$};
	\path (dt.center) + (0.1, -0.3) node [right] {\Huge $d_{t+1}$};
	% Momentum 2
	\path (step-t1.center) + (1,-1.2) node (m-t1-1) {};
	\path [draw, dashed, ->, blue, thick] (step-t1.center)--(m-t1-1.center);
	\path (step-t1.center) + (1.5,0.5) node (m-t1-2) {};
	\path [draw, dashed, ->, blue, thick] (step-t1.center)--(m-t1-2.center);
	\path (step-t1.center) + (2,-0.2) node (m-t1-4) {};
	\path [draw, dashed, ->, blue, thick] (step-t1.center)--(m-t1-4.center);
	\path (m-t1-4.center) + (0.4, -0.04) node (saddle2) {};
	%	\fill [teal!80] (saddle2.center) circle (4pt);
	%	\path (saddle2.center) + (1.8,0.5) node {\huge saddle point};
	% step t+1
	\path (step-t1.center) + (3.4, -1.1) node (step-t2) {};
	\path [draw, red, thick,->] (step-t1.center) -- (step-t2.center) node [midway] (dtt1) {};
	\path (dtt1.center) + (0.4, -0.15) node [below] {\Huge $d_{t+2}$};
	\path (step-t2.center) + (0.30, -0.40) node {\Huge $w_{t+2}$};
	% dot
	\path (step-t2.center) + (1, 0) node (step-t2-1) {};
	\path [draw, dotted, thick, red] (step-t2.center) -- (step-t2-1.center) {};
	% last d
	\path (step-t2-1.center) + (2.7, -1.9) node (last) {} ;
	\path [draw, red, thick, ->] (step-t2-1.center) -- (last.center) node [midway, black, below] {\Huge $d_T$};
	\path (step-t2-1.center) + (0.40, 0.30) node {\Huge $w_{T-1}$};
	\path (last.center) + (-0.40, -0.30) node {\Huge $w_{T}$};
	% legend
	\path (step-t1.center) + (-6.5, -4.8) node (legend) {};
	\path (legend.center) + (0, 0) node(l1h) {};
	\path (legend.center) + (2, 0) node(l1e) {};
	\path [draw, red, thick, ->] (l1h.center) -- (l1e.center) node {};
	\path (l1e.east) + (1.9, 0) node {\huge Update vector};
	\path (legend.center) + (0, -1) node (l2h) {};
	\path (legend.center) + (2, -1) node (l2e) {};
	\path [draw, blue, dashed, thick, ->] (l2h.center) -- (l2e.center) node {};
	\path (l2e.east) + (2.3, 0) node {\huge Candidate vector};
	\path (legend.east) + (8, -0.5) node (l4) {};
	\path (legend.east) + (10, -0.5) node (l3) {};
	% Momentum field
	\path [draw, blue, dashed, thick, ->] (l4.west) -- (l3.east) node  {};
	\path (l3.east) + (2.4, 0) node (legend-field) {\huge Moment field};
	\path [draw, blue, dashed, thick, ->] (l4.west) -- (l1h.north-|l3.east);
	\path [draw, blue, dashed, thick, ->] (l4.west) -- (l2h.south-|l3.east);
	% frameon
	\path (legend.center) + (-0.4, 0.4) node (s1){};
	\path (legend-field.center) + (2.0, -1.0) node (s3){};
	\path [draw, gray] (s1.center)--(s1.center-|s3.center)--(s3.center)--(s3.center-|s1.center)--(s1.center);
	\end{tikzpicture}
	\caption{An illustration of parameter optimization of a learner using proposed HyperAdam algorithm.
	}
	\label{field}
\end{figure}
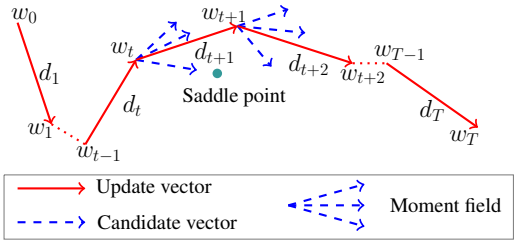

\subsection{General Idea}
Adam is non-adaptive because its hyperparameters (decay rates $\beta$ and $\gamma$ in Alg.~\ref{adam}) are constant and set by hand when optimizing a network. According to Alg.1, different hyperparameters make the parameter updates different both in direction and magnitude. Our proposed HyperAdam improves Adam as follows. First, Adam in HyperAdam is designed with multiple learned task-adaptive decay rates and to generate multiple candidate parameter updates with corresponding decay rates in parallel. Second,  HyperAdam combines these parameter updates to get the final parameter update using adaptively learned combination weights.

As illustrated in Fig. \ref{field}, at a certain point, e.g., $w_t$ in parametric space, multiple update vectors are generated by Adam with different decay rates. The final update $d_t$ is an adaptive combination of these candidate vectors.  Considering that, for a deep neural network \cite{dauphin2014identifying},  there exist abundant saddle points surrounded by high loss plateaus, a certain candidate update vector may point to a saddle point, but an adaptive combination of several candidate vectors may potentially relieve the possibility of getting stuck in saddle point.

\subsection{Task-Adaptive HyperAdam}
Based on the above idea, we design a task-adaptive HyperAdam in Alg. \ref{generalized adam-optimzier}. In iteration $t$, first, the current state $\mathbf{s}_t$ is determined by \textit{state function} $F_h$ with current gradient $g_t$ and previous state $\mathbf{s}_{t-1}$ as inputs in line 8. Then in lines 9-16, $J$ candidate update vectors $\hat{m}_t^j$ are generated by Adam with $J$ pairs of decay rates $(\beta_t^j, \gamma_t^j)$ which are adaptive to the current state $\mathbf{s}_t$ via \textit{decay-rate functions} $F_u$ and $ F_r$. Meanwhile, $J$ task-adaptive weight vectors $\rho_t^j$ are generated by \textit{weight function} $F_q$ with the current state $\mathbf{s}_t$ as input in line 17. Finally, the final update vector $d_t$ is a combination of the candidate updates weighted by  weight vectors in line 18. $\hat{\mathbf{m}}_t$ containing candidate updates and $\boldsymbol{\rho}_t$ containing weight vectors  are called \textit{moment field} and \textit{weight field} respectively.

As illustrated in the left of Fig. \ref{framework},  HyperAdam, as an optimizer,  is a recurrent mapping $O$ iteratively generating  parameter updates. The right of Fig. \ref{framework} shows the graphical diagram of HyperAdam having four components. The StateCell corresponds to the \textit{state function} $F_h$ outputting the current state $\mathbf{s}_t=[s_t^1,\dots,s_t^J]$. With the current state as basis, the \textit{moment field} and \textit{weight field} are produced by AdamCell and WeightCell respectively. The final update $d_t$ is generated in the ``Combination" block. We next introduce these components.

\begin{figure}[!htbp] 
	\centering
	
	\tikzset{global scale/.style={
			scale=#1,
			every node/.append style={scale=#1}
		}
	}
	\begin{tikzpicture}[global scale = 0.56]
	%node
	\node[vectorg](g) {\LARGE$g_t$};
	\path (g.north) + (0, 1.2cm) node [operation2] (norm) {\LARGE Normalization};
	\path (norm.east) + (2.40cm, 0) node [operation2] (pre) {\LARGE Preprocessing};
	\path (pre.east) + (2.0cm, 0) node [operation3] (lstm){\LARGE LSTM};
	\path(lstm.south |- g.center) node [vectorg](s){\LARGE$\mathbf{s}_{t-1}$};
	\path (lstm.east) + (1.6cm, 0cm) node [vectorg](s1){\LARGE$\mathbf{s}_t$};
	% arrow
	\path [draw,->, thick] (g.north) -- (norm.south);
	\path [draw,->, thick] (norm.east)--(pre.west);
	\path [draw,->,thick](pre.east)--(lstm.west);
	\path [draw,->,thick](lstm.east)--(s1.west);
	\path [draw,->,thick](s.north)--(lstm.south);
	% statecell
	\path (norm.west|-norm.north) + (-0.2cm, 0.2cm) node (lu){};
	\path (lstm.east|-lstm.south) + (0.2cm, -0.2cm) node (rb){};
	\path [draw, dashed, thick,] (lu.center) -- (lu.center-|rb.center)--(rb.center)--(lu.center|-rb.center)--(lu.center);  
	\path (norm.north) + (0, 0.5cm) node {\LARGE StateCell};
	\end{tikzpicture}
	
	\caption{Diagram of StateCell. Normalization refers to normalizing the gradient $g_t$ with its Euclidean norm.}
	\label{diag-state}
	
\end{figure}
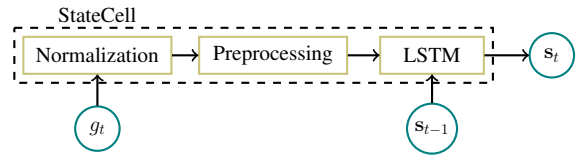

\subsubsection{StateCell}
The current state $\mathbf{s}_t=[s_t^1, \dots, s_t^J]$ is determined by the gradient $g_t$ and previous task state $\mathbf{s}_{t-1}$ in the StateCell implementing the \textit{state function} $F_h$ in line 8 of Alg. \ref{generalized adam-optimzier}.
The diagram of StateCell is illustrated in Fig. \ref{diag-state}. After normalized with its Euclidean norm, the gradient is preprocessed by a fully connected layer with Exponential Linear Unit (ELU) \cite{clevert2016fast} as activation function. Following preprocessing, the gradient together with the previous task state $\mathbf{s}_{t-1}$ are fed into LSTM \cite{lstm1997} to generate the current state $\mathbf{s}_{t}$.
%The dataflow of StateCell is illustrated in Fig. \ref{diag-state}. The gradient after normalization is preprocessed by a fully connected layer with ELU (Exponential Linear Unit) as activation funtion before being handled by LSTMCell. The LSTMCell is also fed with the previous task state $\mathbf{s}_{t-1}$ and outputs the current task state $\mathbf{s}_t$ which is the basement of the middle level. 
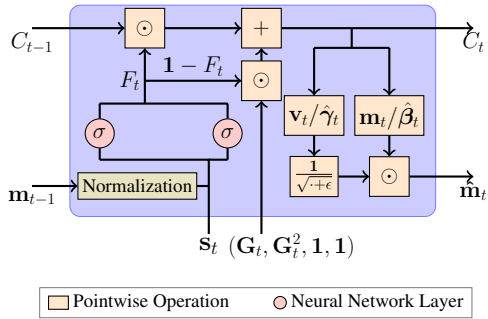
\begin{figure}[!bp] 
	
	\centering
	\tikzset{global scale/.style={
			scale=#1,
			every node/.append style={scale=#1}
		}
	}
	\begin{tikzpicture}[global scale = 0.5] 
	\begin{pgfonlayer}{main}
	\node[cell](cell){};
	\end{pgfonlayer}
	\begin{pgfonlayer}{background}
	\path (cell.west) + (-1.0, 2.20) node [below] (state-in) {\huge$C_{t-1}$};
	\path (cell.east) + (1.0, 2.2) node[below] (state-out) {\huge$C_t$};
	
	\path (cell.west) + (-1.0, -2.00) node [below] (hidenState) {\huge $\mathbf{m}_{t-1}$};
	
	\end{pgfonlayer}
	\begin{pgfonlayer}{foreground}
	\path (state-in.north) + (3, 0) node[operation] (forget) {\huge$\odot$};
	\path [draw,->,thick] (state-in.north) -- (forget.west);
	\path (state-in.north) + (6.1, 0) node[operation] (input) {\huge$+$};
	\path [draw, ->,thick] (forget.east) -- (input.west);
	\path [draw,->,thick] (input.east) -- (state-out.north);
	
	\path (hidenState.north) + (4.7,0) node (concat){};
	
	\path (hidenState.north) + (2.8, 0) node(norm) [norm-adam]{\LARGE Normalization};
	\path[draw,->,thick] (hidenState.north)--(norm.west);
	\path[draw,-,thick] (norm.east)--(concat.center);
	
	\path (concat.center) + (0, -1.4) node (grad) {};
	\path (grad)+ (0, -0.2cm) node {\Huge${\mathbf{s}}_t$};
	\path [draw,-,thick] (grad.north) -- (concat.center);
	
	\path (concat.center) + (0, 0.7) node (split){};
	\path (split.center) + (-2.9, 0) node (s1){};
	\path (split.center) + (0.5, 0) node (s2){};
	\path [draw, thick](concat.center)--(split.center)--(s1.center) (split.center)--(s2.center);
	\path (s1.center) + (0, 0.7) node [nonlinear](sigmoid1) {\huge$\sigma$};
	\path (s2.center) + (0, 0.7) node [nonlinear](sigmoid2) {\huge$\sigma$};
	\path [draw,thick] (s1.center)--(sigmoid1.south) (s2.center)--(sigmoid2.south);
	\path (sigmoid1.north)+ (0, 0.4) node (s3){};
	\path (sigmoid2.north) + (0,0.4) node (s4){};
	\path[draw, thick] (sigmoid1.north) -- (s3.center) (sigmoid2.north)--(s4.center) (s3.center)--(s4.center);
 	\path [draw, thick,->](s3.center-| forget.center)--(forget.south);

	\path (input.center)+(0, -1.4) node[operation] (gate) {\huge$\odot$};
	\path [draw, ->,thick] (gate.north) -- (input.south);
	\path (gate.center -| forget.center) node [left](split) {\huge$F_t$};
	\path [draw, ->,thick](split.east) -- (gate.west) node[midway,above] {\huge$\mathbf{1}-F_t$};

	\path [draw, ->,thick](grad.north -| gate.center) --(gate.south);
	\path (grad.center -| gate.center) + (0.8cm, -0.2cm) node{\huge$(\mathbf{G}_t, \mathbf{G}^2_t, \mathbf{1}, \mathbf{1})$};

	\path (state-in.north) + (8.5, 0) node (out) {};
	\path (out.south) + (0, -0.4) node (out-s) {};
	\path [draw,-,thick] (out.center) -- (out-s.center);
	
	\path (out-s.center) + (-1, 0) node (v) {};
	\path (out-s.center) + (1,0) node (m) {};
	\path [draw,-,thick] (m.center) -- (v.center);
	
	\path (v.center) + (0, -1.8) node[operation] (vc){\huge${\mathbf{v}_t}/{\hat{\boldsymbol{\gamma}}_{t}}$};%\huge$\frac{\mathbf{v}_t}{\hat{\boldsymbol{\gamma}}_{t}}$
	\path (m.center|-vc.center) node[operation]
	(mc){\huge${\mathbf{m}_t}/{\hat{\boldsymbol{\beta}}_{t}}$};%{$\frac{m_t}{\hat{\beta}_{(m,t)}}$};
	
	\path [draw,->,thick] (v.center) --(vc.north);
	\path [draw,->,thick] (m.center) -- (mc.north);
	
	\path (v.south) + (0, -3.3) node [operation] (daoshu)
	{\huge$\frac{\mathbf{1}}{\sqrt{\cdot + \epsilon}}$};
	% {$\frac{1}{\sqrt{\cdot + \epsilon}}$};
	
	\path[draw,->,thick] (vc.south) -- (daoshu.north);
	
	\path (daoshu.center -| m.center) node[operation] (xiangcheng) {\huge$\odot$};
	\path [draw, ->,thick] (mc.south) -- (xiangcheng.north) ;
	\path [draw, ->,thick] (daoshu.east) -- (xiangcheng.west);
	
	\path (xiangcheng.center -| cell.east) + (1, 0) node [below] (out) {\huge$\mathbf{\hat{m}}_{t}$};
	\path [draw,->,thick] (xiangcheng.east) -- (out.north);
	\end{pgfonlayer}
	\path (hidenState.south) + (2em, -7em) node[operation-legend] (legend1){};
	\path (legend1.east) node [right] {\LARGE Pointwise Operation};
	\path (legend1.east) + (16em, 0) node[nonlinear] (legend2){};
	\path (legend2.east) node [right] {\LARGE Neural Network Layer};
	% frameon
	\path (legend1.center) + (-0.5cm, 0.4cm) node (ss11){};
	\path (legend2.center) + (5.1cm,-0.4cm) node (ss33){};
	\path [draw, gray] (ss11.center)--(ss11.center-|ss33.center)--(ss33.center)--(ss33.center-|ss11.center)--(ss11.center);
	\end{tikzpicture}
	
	\caption{Diagram of AdamCell. The neural network layer corresponds to the decay-rate functions $F_u,F_r$ in Alg. \ref{generalized adam-optimzier}.}
	\label{AdamCell}
	
\end{figure}

\subsubsection{AdamCell}
AdamCell is designed to implement lines 9-16 in Alg. \ref{generalized adam-optimzier} for generating moment field, i.e. a group of update vectors. We first analyze these lines. Task-adaptive decay rates $\boldsymbol{\beta}_t, \boldsymbol{\gamma}_t$ are generated by \textit{decay-rate functions} in lines 9-10, with which the biased estimations of first and second moment of gradients $\mathbf{m}_t, \mathbf{v}_t$ are recursively accumulated in lines 11-12. The bias factors $\hat{\boldsymbol{\beta}}_{t}, \hat{\boldsymbol{\gamma}}_{t}$ are computed in lines 13-14. Finally, moment field is produced with unbiased estimations of first and second moment of gradients in line 16. 

Note that the  accumulated $\hat{\boldsymbol{\beta}}_{t}$ (line 13 of Alg.~\ref{generalized adam-optimzier}) is equivalent to $\mathbf{1}-\boldsymbol{\beta}^t$ when $\boldsymbol{\beta}$ is constant based on lemma \ref{online}. It also holds for $\hat{\boldsymbol{\gamma}}_{t}$. Therefore, we can derive that each component in moment field (line 16 of Alg. \ref{generalized adam-optimzier}) is equivalent to a parameter update produced by Adam in line 9 of Alg.~\ref{adam}.
\begin{lemma}\label{online}
	$\hat{\beta}_t = \beta \hat{\beta}_{t-1} + (1-\beta)$ with $\hat{\beta}_0=0$ is the online formula of $\hat{\beta}_t=1-\beta^t$.
\end{lemma}
\begin{lemmaproof}
	See proof in supplementary material.
\end{lemmaproof}

If we denote $C_t=[\mathbf{m}_{t},\mathbf{v}_{t}, \hat{\boldsymbol{\beta}}_{t}, \hat{\boldsymbol{\gamma}}_{t}]$, $F_t=[\boldsymbol{\beta}_t, \boldsymbol{\gamma}_t, \boldsymbol{\beta}_t, \boldsymbol{\gamma}_t]$ and $\tilde{C}_t=[\mathbf{G}_t, \mathbf{G}^2_t, \mathbf{1}, \mathbf{1}]$ ($\mathbf{1}\in\mathbb{R}^{p\times J}$), based on  lemma \ref{online}, lines 11-14 in Alg. \ref{generalized adam-optimzier}  can be expressed in the following compact formula resembling cell state updating in LSTM:
\begin{equation} \label{state-function}
C_t = F_t \odot C_{t-1} + (\mathbf{1}-F_t) \odot \tilde{C}_t.
\end{equation}
 Thus we construct AdamCell, a structure like LSTM, to conduct lines 9-16 in Alg.~\ref{generalized adam-optimzier} as illustrated in Fig. \ref{AdamCell}. $F_t$ determines how much historical information would be forgot like the forget gate in LSTM. We define the \textit{decay-rate functions} $F_r, F_u$ in Alg.~\ref{generalized adam-optimzier} to be in parametric forms:
\begin{align}
	\boldsymbol{\beta}_t &= \sigma([\mathbf{m}_{t-1}',\mathbf{s}_t]\boldsymbol{\theta}_{u} + \mathbf{b}_{u}), \label{beta}  \\
	 \boldsymbol{\gamma}_t &= \sigma([\mathbf{m}_{t-1}',\mathbf{s}_t]\boldsymbol{\theta}_{r} + \mathbf{b}_{r}), \label{gamma}
\end{align}
with $\boldsymbol{\theta}_{u}, \boldsymbol{\theta}_{r}\in\mathbb{R}^{2J\times J}$, $\mathbf{b}_{u}=[b_u, \dots, b_u]^T \in \mathbb{R}^{p\times J}$,  $\mathbf{b}_{r}=[b_r, \dots, b_r]^T \in \mathbb{R}^{p\times J}$ and $\mathbf{m}_{t-1}' = [\frac{m_t^1}{\lVert m_t^1 \rVert_2}, \dots, \frac{m_t^J}{\lVert m_t^J \rVert_2}]\in \mathbb{R}^{p\times J}$, where $\varTheta_u = \{\boldsymbol{\theta}_{u}, {b}_{u}\}, \varTheta_r= \{\boldsymbol{\theta}_{r}, {b}_{r}\}$ are learnable parameters. The decay-rate functions $F_r$, $F_u$ output decay rates $\boldsymbol{\beta}_t=[\beta_t^1, \dots\beta_t^J]$ and $\boldsymbol{\gamma}_t=[\gamma_t^1, \dots\gamma_t^J]$ respectively, and each pair of decay rates $(\beta_{t}^j, \gamma_{t}^j)$ determines a candidate update vector $\hat{m}_t^j$ generated by Adam. 

\subsubsection{WeightCell}
WeightCell is designed to implement the \textit{weight function} $F_q$ (line 17 in Alg.~\ref{generalized adam-optimzier}) which outputs the \textit{weight field} with the current state $\mathbf{s}_t$ as input. The weight function is a one-hidden-layer fully connected network with ELU as activation function:
\begin{equation}
	\boldsymbol{\rho}_t = \text{ELU}(\mathbf{s}_t\boldsymbol{\theta}_q+\mathbf{b}_q),
\end{equation} 
with $\boldsymbol{\theta}_{q} \in \mathbb{R}^{J\times J}$ and $\mathbf{b}_{q}=[b_q, \dots, b_q]^T \in \mathbb{R}^{p\times J}$ where $\varTheta_q=\{ \boldsymbol{\theta}_q, b_q \}$ are learnable parameters. 

We choose ELU instead of ReLU as activation function to ensure that the weights are not always positive, since some candidate vectors in the \textit{moment field} may not be favorable because of pointing to a bad direction.
\subsubsection{Combination}
The final update $d_t$ is the combination of the candidate update vectors in moment field with weight vectors in weight field (line 18 in Alg.~\ref{generalized adam-optimzier}):
\begin{equation}
	d_t = \Sigma_{j=1}^J \rho_{t}^j \odot \hat{m}_t^j.
\end{equation}
\subsubsection{Parameter sharing} 
It can be verified that the different coordinates of parameter $w$ and intermediate terms such as $\mathbf{s}_t, \boldsymbol{\beta}_t, \boldsymbol{\gamma}_t$ share the hyperparameter $\varTheta=\{\varTheta_h, \varTheta_q, \varTheta_r, \varTheta_u\}$ of HyperAdam. For example, different rows of $\boldsymbol{\beta}_t$ in Eqn.~(\ref{beta}), corresponding to different coordinates of $w$, share hyperparameters $\varTheta_u$. Moreover, $J$ candidate update vectors are generated in parallel by matrix operations. Consequently, HyperAdam can be applied to training networks with varying dimensional parameters in parallel. 
\subsubsection{Scale invariance}
To achieve the scale invariance  property same as traditional Adam, the gradient $g_t$ and $m_t^j$ ($j=1,\dots,J$)  are normalized by their Euclidean norms in StateCell and AdamCell (see proof in supplementary material).
\section{Learning HyperAdam} % \footnote{\url{https://www.tensorflow.org}}
We train HyperAdam on a \textit{meta-train} set consisting of learner (i.e., network in this paper) coupled with corresponding optimizee (loss for training learner) and dataset, which is implemented by TensorFlow. We aim to optimize the parameters of HyperAdam to maximize its capability in training learners over the meta-train set. We expect that the learned HyperAdam can be generalized to optimize more complex networks beyond the learners in meta-train set.  We next introduce the training process in details.

Meta-train set consists of triplets of learner $f$, optimizee $L$, and dataset  $\mathcal{D}=\{ X,Y \}$, where $X=\{x_i\}_{i=1}^N$  and $Y=\{y_i\}_{i=1}^N$ represent the data set and corresponding label set. The HyperAdam parameter set $\varTheta$ is optimized by minimizing  the expected cumulative regret~\cite{andrychowicz2016learning} on the meta-train set:
\begin{equation}
\mathcal{L}(\varTheta) = \mathbb{E}_{L}[\frac{1}{T}\Sigma_{t=1}^T L(f (X;w_t(\varTheta)),Y)],
\end{equation}
where $w_{t}(\varTheta) = w_{t - 1}(\varTheta) -\alpha d_{t} (g_{t}, \varTheta)$ is  network parameter of learner $f$ at iteration $t$ when optimized by HyperAdam with parameters $\varTheta=\{\varTheta_h,\varTheta_q,\varTheta_r,\varTheta_u\}$. $f (X;w_t(\varTheta))$ denotes the network output of learner $f$ on dataset $\mathcal{D}$ when network parameter is $w_t(\varTheta)$, and $L(\cdot,\cdot)$ is an optimizee, i.e., the loss for training learner $f$. Therefore, $\mathcal{L}(\varTheta)$ defines the expectation of  the cumulative loss over meta-train set. Minimizing $\mathcal{L}(\varTheta)$ is to find optimal parameter for HyperAdam to reduce training loss $L$ (i.e., optimizee) as lower as possible.

As in~\cite{lv2017learning}, the learner $f$ is simply taken as a forward neural network with one hidden layer of 20 units and sigmoid as activation function. The optimizee $L$ is defined as  $L(f (X;w),Y)=\Sigma_{i=1}^N l(f(x_i;w),y_i)$ where $l$ is the cross entropy loss for the learner $f$ with a minibatch of 128 random images sampled from the MNIST dataset \cite{lecun1998gradient}.  We set the learning rate  $\alpha =0.005$ and maximal iteration $T=100$ indicating the number of optimization steps using HyperAdam as an optimizer. The number of candidate updates $J$ is set to be 20. 

HyperAdam can be seen as a recurrent neural network iteratively updating network parameters. Therefore we can optimize parameter $\varTheta$ of HyperAdam using  BackPropagation Through Time  \cite{werbos1990bptt} by minimizing $\mathcal{L(\varTheta)}$ with Adam, and the expectation with respect to $L$ is approximated by the average training loss for learner $f$ with different initializations. The $T=100$ steps are split into 5 periods of 20 steps to avoid gradient vanishing. In each period, the initial parameter $w_0$ and initial hidden state $\mathcal{H}$ are initialized from the last period or generated if it is the first period.
% \footnote{HyperAdam is optimized using BPTT with Adam solver.}
%HyperAdam can be seen as a recurrent neural network iteratively updating network parameters. Therefore we can optimize parameter $\varTheta$ of HyperAdam by minimizing $\mathcal{L(\varTheta)}$ with Adam, and the expectation with respect to $L$ is approximated by the average training loss for learner $f$ with different initializations. The $T=100$ steps are split into 5 periods of 20 steps to avoid gradient vanishing. In each period, the initial parameter $w_0$ and initial hidden state $\mathcal{H}$ are initialized from the last period or generated if it is the first period.

Two training tricks proposed in \cite{lv2017learning} are used here. First, in order to make the training easier, a $k$-dimensional convex function  $h(z)=\frac{1}{k}\lVert z- \eta \rVert^2$ is combined with the original optimizee (i.e., training loss), and this trick is called ``\textbf{Combination with Convex Function}" (CC).  $\eta$ and initial value of $z$  are generated randomly. Second, ``\textbf{Random Scaling}'' (RS), helping to avoid over-fitting, randomly samples vectors $c_1$ and $c_2$ of the same dimension as parameter $w$ and $z$ respectively, and then multiply the parameters with $c_1$ and $c_2$ coordinate-wisely, thus the optimizee in the meta-train set becomes:    
\begin{equation}
L_{ext}(w, z) = L(f(X;c_1\odot w),Y) + h(c_2 \odot z),
\end{equation} 
with initial parameters $diag(c_1)^{-1}w, diag(c_2)^{-1}z$.
\begin{figure*}[!htbp]
	\centering
	
	\includegraphics[scale=0.27]{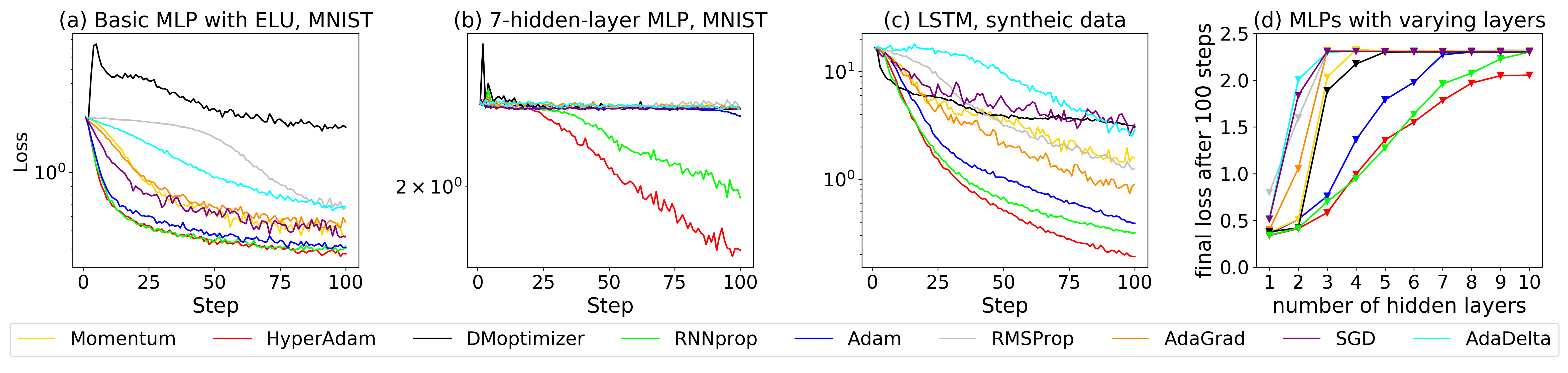}

	\caption{HyperAdam performs best compared with other optimizers on neural networks with different structures.}
	\label{elu-l7-lstm-deep}
	
\end{figure*}
\begin{figure*}[!htb]
	
	\centering
	\includegraphics[scale=0.27]{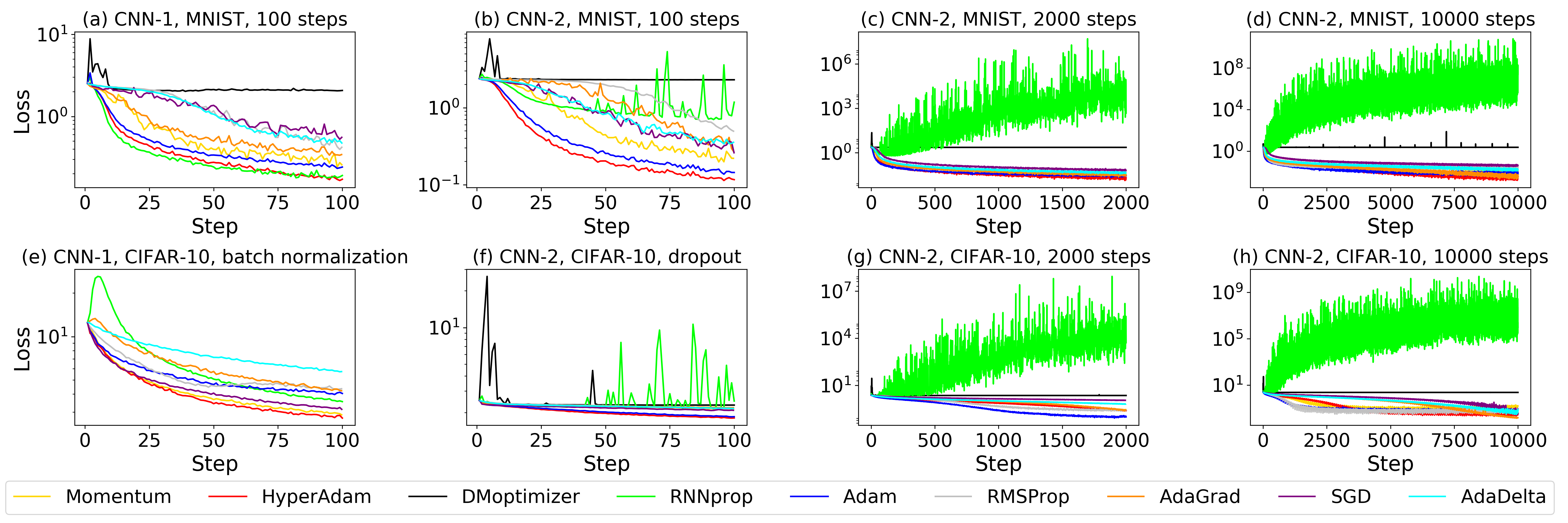}

	\caption{Comparison of different optimizers for optimizing different CNNs in different optimization steps. 
	}
	\label{cnnallreg}	
	
\end{figure*}

\section{Evaluation}
We have trained HyperAdam based on 1-layer  MLP (basic MLP), we now evaluate the learned HyperAdam for more complex networks such as basic MLP with different activation functions, deeper MLP, CNN and LSTM.
\begin{itemize}
	\item \textbf{Activation functions:} The activation function of  basic MLP is extended from sigmoid to ReLU, ELU and tanh.
	\item \textbf{Deep MLP:} The number of hidden layers of MLP is extended to range of  $[2,10]$, and each layer has 20 hidden units and uses sigmoid as activation function.
	\item \textbf{CNN:} Convolution neural networks are with structures of   $c$-$c$-$p$-$f$ (CNN-1) and $c$-$c$-$p$-$c$-$c$-$p$-$f$-$f$ (CNN-2), where  $c$, $p$ and $f$ represent  convolution, max-pooling and fully-connected layer respectively. Convolution kernel is with size of $3\times 3$ and the max-pooling layer is with size of  $2\times2$ and stride 2. CNN-1 and CNN-2 are also trained with batch normalization and dropout respectively.
	\item \textbf{LSTM:} LSTM with hidden state in size of 20 is applied to sequence prediction task using mean squared error loss as in \cite{lv2017learning}. Given a sequence $f(0), \dots, f(9)$ with additive noise, the LSTM is supposed to predict the value of $f(10)$. Here $f(x) = A\sin(wx+\phi)$. The dataset is generated with uniformly random sampling $A \sim U(0, 10), w\sim U(0, \pi/2),\phi\sim U(0, 2\pi)$, and the noise is drawn from Gaussian distribution $N(0, 0.1)$.
\end{itemize}
We also evaluate whether our learned HyperAdam can well generalize to different datasets, e.g. CIFAR-10 \cite{krizhevsky2009learning}. Moreover, the HyperAdam is trained assuming it iteratively optimizes network parameters in fixed iterations $T=100$, we also evaluate the learned HyperAdam for longer iterative optimization steps as in~\cite{lv2017learning}. The generalization ability of the networks trained by HyperAdam will be also evaluated preliminarily.

In evaluations, we will compare our HyperAdam with traditional network optimizers such as SGD, AdaDelta, Adam, AdaGrad, Momentum, RMSProp, and state-of-the-art learning-based optimizers including RNNprop \cite{lv2017learning}, DMoptimizer \cite{andrychowicz2016learning}. For the traditional optimizers, we hand-tuned the learning rates and set other hyperparameters as defaults in TensorFlow. All the initial parameters of learners used in the experiments are sampled independently from the Gaussian distribution. We report the quantitative value as  the average measure for training the learner 100 times with random parameter initialization.

\subsection{Generalization with Fixed Optimization Steps}
We first assume that the learned HyperAdam optimizes the parameters of learners for fixed optimization steps $T = 100$, same as the learning procedure for HyperAdam. 
\begin{table}[!bp]
	\centering
	\scalebox{0.75}{
		\begin{tabular}{c|cccc}
			\hline
			Activation&Adam&DMoptimizer&RNNprop&HyperAdam \\
			\hline
			sigmoid&0.35 & 0.38 &0.34 &\textbf{0.33} \\
			%\midrule
			ReLU & 0.32& 1.42&0.31& \textbf{0.29}\\
			%\midrule
			ELU&0.31&2.02&0.31&\textbf{0.28}\\
			%\midrule
			tanh&0.34&0.83&\textbf{0.33}&0.36\\
			\hline
		\end{tabular}	
	}
	
	\caption{Performance for training basic MLP in 100 steps with different activation functions. Each value is the average final loss for optimizing networks in 100 times.}
	\label{activations}
	
\end{table}
\subsubsection{Activation functions} 
As shown in Table \ref{activations}, HyperAdam is tested for training basic MLP with different activation functions on MNIST dataset, the loss values in Table \ref{activations}  show that HyperAdam can best generalize to optimize basic MLP with ReLU, ELU and tanh as activation functions, compared with DMoptimizer and RNNprop. Our HyperAdam also outperforms the basic Adam algorithm. The DMoptimizer can not well generalize to basic MLP with ELU activation function, which can be also visually observed in Fig. \ref{elu-l7-lstm-deep}(a).

\subsubsection{Deep MLP} We further evaluate performance of HyperAdam on learning parameters of  MLPs with varying layer numbers.  According to Fig. \ref{elu-l7-lstm-deep}(d), for different number of hidden layers ranging from 1 to 10,  HyperAdam always performs significantly better than Adam and DMoptimizer. Compared with RNNprop, HyperAdam is better in general, especially for deeper MLP with more than 6 layers. The loss curves in Fig. \ref{elu-l7-lstm-deep}(b) of different optimizers for MLP with 7 hidden layers illustrate HyperAdam is significantly better.

\subsubsection{LSTM} 
As shown in Table \ref{lstm}, the ``Baseline'' task is to utilize one-layer LSTM to predict $f(10)$ on dataset with noise drawn from $N(0, 0.1)$, which is further varied by training on dataset with small noise drawn from $N(0, 0.01)$ (``Small noise'') or using two-layer LSTM (``2-layer'') for prediction. By comparing the loss values in Table \ref{lstm}, our HyperAdam can better decrease the training losses than the compared optimizers, i.e., Adam, DMoptimizer, RNNprop, HyperAdam. Specifically, Fig. \ref{elu-l7-lstm-deep}(c) shows an example for the comparison in task of ``Small noise''. 

\begin{figure}[!hb]
	
	\centering
	\includegraphics[scale=0.345]{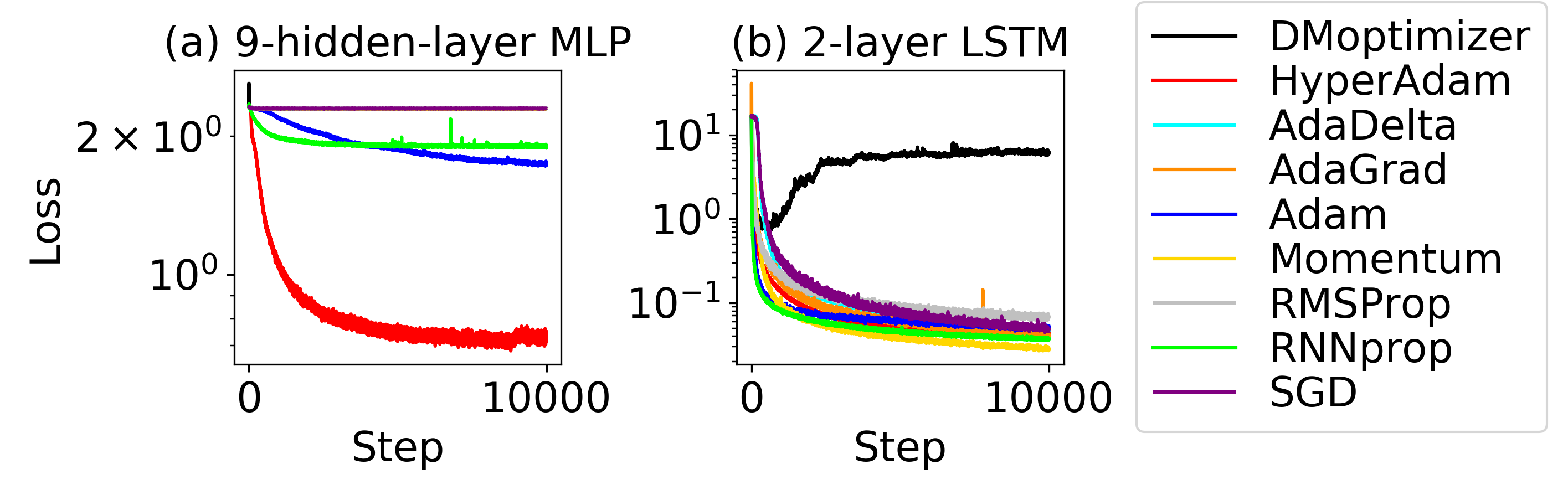}
	
	\caption{HyperAdam with 10000 optimization steps. Training curves by DMoptimizer, AdaGrad, RMSProp, AdaDelta, Momentum and SGD coincide in the left figure.}
	
	\label{l9-lstm-longer}	
\end{figure}
\begin{table}[!htbp]
	\centering
	
	\scalebox{0.75}{
		\begin{tabular}{c|cccc}
			\hline
			Task&Adam&DMoptimizer&RNNprop&HyperAdam \\
			\hline
			Baseline&0.65 & 3.10 &0.49 &\textbf{0.42} \\
			Small noise&0.39&3.06&0.32&\textbf{0.19}\\
			%\midrule
			2-layer & 0.51& 2.05&0.27& \textbf{0.26}\\
			%\midrule
			\hline
		\end{tabular}	
	}
	\caption{Performance on different sequence prediction tasks. }
	\label{lstm}
\end{table}
\subsubsection{CNN} 	
 Figure \ref{cnnallreg}(a)-(b) compare  training curves of CNN-1  and CNN-2 on MNIST using different optimizers.  Figure \ref{cnnallreg}(e)-(f) compare training curves of CNN-1 with batch normalization and CNN-2 with dropout on CIFAR-10 respectively. In these figures, DMoptimizer and RNNprop do not always perform well or even fail while HyperAdam can effectively decrease the training losses in different tasks.

 \begin{figure*}[!htb]
 	
 	\centering
 	\includegraphics[scale=0.27]{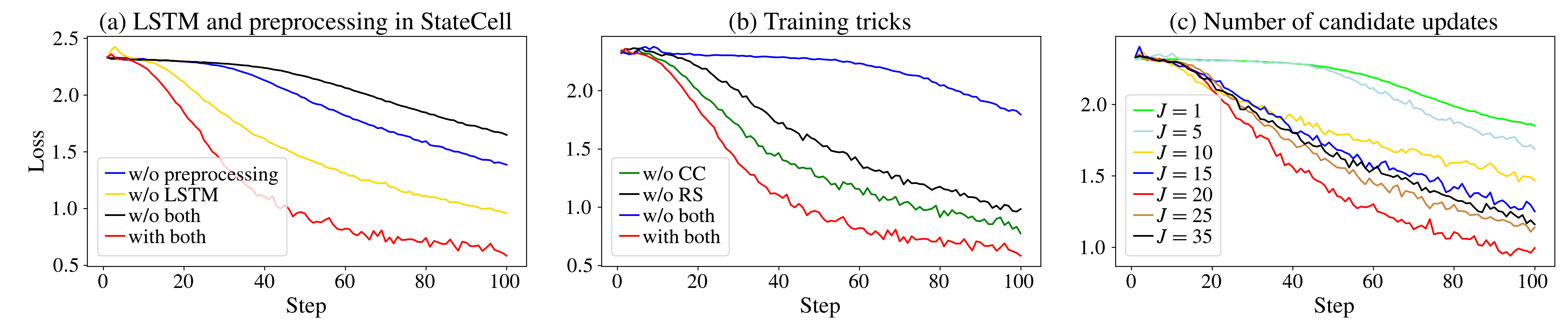}
 	
 	\caption{Ablation study for {the number of candidate updates}, {training tricks} and {structure in StateCell}. CC denotes ``Combination with Convex function", and RS denotes ``Random Scaling". }
 	\label{ablation}	
 \end{figure*}
\subsection{Generalization to Longer Horizons}
We have evaluated HyperAdam for optimizing different learners in fixed optimization steps ($T =100$), same as the meta-training phase. We now evaluate HyperAdam for its effectiveness in running optimization for longer steps. 

\subsubsection{Deep MLP}
%As shown in Fig. \ref{l9-lstm-longer}(a), HyperAdam performs significantly better than other optimizers on training MLP with 9 hidden layers in 10000 steps.  Adam and RNNprop decrease the loss  slowly and other optimizers all fail, but HyperAdam achieves a much lower training loss. 
Figure \ref{l9-lstm-longer}(a) illustrates the training curves of MLP with 9 hidden layers on MNIST using different optimizers for 10000 steps. DMoptimizer and almost all traditional optimizers, including SGD, Momentum, AdaGrad, AdaDelta and RMSProp, fail to decrease the loss. Our HyperAdam can effectively decrease the training loss.
\subsubsection{LSTM}
The comparison of training two-layer LSTM to predict $f(10)$ with different optimizers for 10000 steps is shown in Fig. \ref{l9-lstm-longer}(b). DMoptimizer decreases the loss first and then increases the loss dramatically. With similar performance to the traditional optimizers, such as AdaGrad and AdaDelta, our HyperAdam and RNNprop perform better than RMSProp and SGD.
 
\subsubsection{CNN}
Figure \ref{cnnallreg}(c)-(d) show the training curves of CNN-2 for 2000 steps and 10000 steps on MNIST dataset. Both RNNprop and  DMoptimizer fail to decrease the loss. However, HyperAdam manages to decrease the loss of CNNs and performs slightly better than the traditional network optimizers, such as SGD, Adam and AdaDelta.  When training CNN-2 on CIFAR-10 dataset, HyperAdam does not perform as fast as Adam and RMSProp for the first 2000 steps according to Fig. \ref{cnnallreg}(g), but achieves lower loss at 10000th step as shown in Fig. \ref{cnnallreg}(h), while RNNprop and DMoptimizer fail to sufficiently decrease the training loss.

\subsection{Generalization of the Learners}
The generalization ability of the learners trained by DMoptimizer, RNNprop, Adam and HyperAdam for 10000 steps is evaluated. Table \ref{testLearner} shows the loss, top-1 error and top-2 error of the two learners, CNN-1 and CNN-2 on dataset MNIST, which shows the generalization of learners trained by HyperAdam and Adam are significantly better than those trained by DMoptimizer and RNNprop. 
\subsection{Ablation Study}
We next perform ablation study to justify  the  effectiveness of   key components in HyperAdam.
\begin{table}[!htbp]
	\centering
	\scalebox{0.67}{
		\begin{tabular}{c|c|cccc}
			\hline
			Task&Measure &Adam&DMoptimizer&RNNprop&HyperAdam \\
			\hline
			\multirow{3}*{\parbox{6.7em}{\centering CNN-1\\ (MNIST)}} & loss &0.10&2.30&0.36&\textbf{0.05}\\
			& top-1&\textbf{98.50}\%&10.10\%&96.46\%&{98.48\%}\\
			& top-2&99.59\%&20.38\%&99.03\%&\textbf{99.63\%}\\
			\hline
			\multirow{3}*{\parbox{6.7em}{\centering CNN-2\\ (MNIST)}} & loss &0.09&2.30&2.30&\textbf{0.07}\\
			& top-1&98.98\%&11.35\%&11.37\%&\textbf{99.02\%}\\
			& top-2&\textbf{99.80}\%&21.45\%&21.69\%&{99.78\%}\\
			\hline
		\end{tabular}	
	}
	
	\caption{Generalization of the learner trained by Adam, DMoptimizer, RNNprop and HyperAdam for 10000 steps.}
	\label{testLearner}
	
\end{table}

\subsubsection{LSTM and preprocessing in StateCell} Figure \ref{ablation}(a) illustrates that HyperAdam achieves lower loss than HyperAdam without LSTM block or (and) preprocessing for training 3-hidden-layer MLP on MNIST, which reflects that the LSTM block and preprocessing help strengthen HyperAdam.

\subsubsection{Training tricks} We justify the effectiveness of ``Random Sampling'' and ``Combination with Convex Functions'' in our proposed HyperAdam.  As shown in Fig. \ref{ablation}(b), HyperAdam trained with both two tricks performs better than HyperAdam trained with either one of them and neither of them for training loss of 3-hidden-layer MLP on MNIST as optimizee, which indicates that the two tricks can enhance the generalization ability of learned HyperAdam.

\subsubsection{Number of candidate updates} Figure \ref{ablation}(c) shows the comparison for optimizing cross entropy loss of 4-hidden-layer MLP on MNIST dataset with HyperAdam having different number of candidate updates ($J=1,5,10,15,20,25,35$). It is observed that the performance of HyperAdam is improved first with the increase of $J$ until 20 achieving best performance, then becomes saturated and decreased with larger number of candidate updates. But all the HyperAdams with $J = 5, 10, 15, 25, 35$ are better than the baseline with single candidate update. 

\subsection{Computation Time}
The time consuming for computing each update by HyperAdam is roughly the same with that of DMoptimizer and RNNprop. For example, the time consuming for computing each update given gradient of a 9-hidden-layer MLP is 0.0023s, 0.0033s and 0.0039s for DMoptimizer, HyperAdam and RNNprop respectively in average. Though faster for computing each update than HyperAdam, Adam is not as efficient as HyperAdam to sufficiently decrease the training loss. When training 8-hidden-layer MLP for 10000 steps, HyperAdam takes 26.33s to decrease the loss to 0.64 (the lowest loss that Adam can achieve) while Adam takes 28.97s to decrease loss to 0.64 and finally achieves loss of 0.52.

\section{Conclusion and Future Work}
In this paper, we proposed a novel optimizer HyperAdam implementing ``learning to optimize'' inspired by the  traditional Adam optimizer and ensemble learning. It  adaptively combines the multiple candidate parameter updates generated by Adam with multiple adaptively learned decay rates. Based on this motivation, a carefully designed RNN was proposed for implementing HyperAdam optimizer. It was justified to outperform or match traditional optimizers such as Adam, SGD and state-of-art learning-based optimizers in diverse networks training tasks. 

In the future, we are interested in applying HyperAdam to train larger scale and more complex networks in vision, NLP, etc., and modeling the correlations among parameter coordinates to further enhance its  performance.
\section{Acknowledgment}
This research was supported by the National Natural Science Foundation of China under Grant Nos. 11622106, 11690011, 61721002, 61472313.
\bibliographystyle{aaai}
\bibliography{paper4964}
\section{Supplementation}
In this supplementary material, we will prove lemma 1 and the claim that HyperAdam is invariant to the scale of gradient (scale invariance) in the submitted paper.

We first present the proof of lemma 1 in Section 3.2.

\begin{lemmas}
	$\hat{\beta}_t = \beta \hat{\beta}_{t-1} + (1-\beta)$ with $\hat{\beta}_0=0$ is the online formula of $\hat{\beta}_t=1-\beta^t$.
\end{lemmas}
\begin{lemmaproofs} Since $\hat{\beta}_0=0$, we derive that:
	\begin{align*}
	\hat{\beta}_t &= \beta \hat{\beta}_{t-1} + (1-\beta) \notag\\
	&=\beta(\beta \hat{\beta}_{t-2} + (1-\beta) ) + (1-\beta)\notag\\
	&=\cdots \notag\\
	&=\beta^t\hat{\beta}_0 + (1-\beta)\Sigma_{i=1}^t\beta^{t-i}\notag\\
	&=1-\beta^t, 
	\end{align*}
	so the lemma is proved. \hfill $\blacksquare$
\end{lemmaproofs}

We then prove the claim on \textbf{scale invariance} described in Section 3.2.

\begin{claim}
	Same to Adam, HyperAdam is also invariant to the scale of gradient when ignoring $\varepsilon$.
\end{claim}
\begin{claimproof}
	In order to prove the claim, we only need to verify that $\mathbf{s}_t, \boldsymbol{\beta}_t, \boldsymbol{\gamma}_t$ and $\boldsymbol{\rho}_t$ are invariant to the scale of gradient. 
	
	Since the gradient will be normalized when it acts as input of the state function $F_h$ corresponding to StateCell, we can rewrite $F_h$ as
	\begin{equation}
	\mathbf{s}_t = F_h(\mathbf{s}_{t-1},g_t;\varTheta_h) \triangleq F'_h(\mathbf{s}_{t-1},\frac{g_t}{\lVert g_t\rVert};\varTheta_h).
	\end{equation}
	We next consider the scaled gradient $sg_t$:
	\begin{align*}
	F_h(\mathbf{s}_{t-1},sg_t;\varTheta_h) &= F'_h(\mathbf{s}_{t-1},\frac{sg_t}{\lVert sg_t\rVert};\varTheta_h) \notag \\
	&= F'_h(\mathbf{s}_{t-1},\frac{g_t}{\lVert g_t\rVert};\varTheta_h) \notag \\
	&=F_h(\mathbf{s}_{t-1},g_t;\varTheta_h).
	\end{align*}
	so $\mathbf{s}_t$ outputted by StateCell is invariant to the scale of the gradient. We can similarly derive  that $\boldsymbol{\rho}_t=F_q(\mathbf{s}_t;\varTheta_q)$, corresponding to the WeightCell, is invariant to the scale of gradient.  Furthermore, the fact that $\boldsymbol{\beta}_t, \boldsymbol{\gamma}_t$ are invariant to the scale of gradient can also be proved in the same way.
	
	Since  decay rates $\boldsymbol{\beta}_t, \boldsymbol{\gamma}_t$ are scale invariant, candidate updates $\hat{m}_t^j$ ($j=1,\dots,J$) outputted by AdamCell are invariant to the scale of gradient. Therefore, the final update $d_t=\Sigma_{j=1}^J \rho_{t}^j \odot \hat{m}_t^j$ is invariant to the scale of gradient.
	\hfill $\blacksquare$
\end{claimproof}
\end{document}